\documentclass[lettersize,journal]{IEEEtran}
\usepackage{amsmath,amsfonts}
\usepackage{algorithmic}
\usepackage{algorithm}
\usepackage{array}
\usepackage[caption=false,font=normalsize,labelfont=sf,textfont=sf]{subfig}
\usepackage{textcomp}
\usepackage{stfloats}
\usepackage{url}
\usepackage{verbatim}
\usepackage{graphicx}
\usepackage{cite}
\usepackage{booktabs}
\usepackage{multirow}
\usepackage{bm}

\begin{document}

\title{Knowledge-Informed Multi-Agent Trajectory Prediction at Signalized Intersections for Infrastructure-to-Everything}

\author{Huilin Yin, Yangwenhui Xu, Jiaxiang Li, Hao Zhang, Gerhard Rigoll~\IEEEmembership{Fellow,~IEEE}
\thanks{Manuscript received XX xx, 2025; revised xx xx, 2025. This work was supported by the National Natural Science Foundation of China under Grant No.62433014. }
\thanks{Huilin Yin, Yangwenhui Xu, Jiaxiang Li are with the
College of Electronic and Information Engineering, Tongji University, Shanghai 201804, China (e-mail:yinhuilin@tongji.edu.cn, xuyangwenhui@tongji.edu.cn,
2230747@tongji.edu.cn).}
\thanks{Hao Zhang is with Shanghai Institute of Intelligent Science and Technology, Tongji University, Shanghai 200092, China, and also with the
College of Electronic and Information Engineering, Tongji University, Shanghai 201804, China
(zhang\underline{~}hao@tongji.edu.cn).}
\thanks{Gerhard Rigoll is with the Department of Electrical and Computer
Engineering, Technical University of Munich, 80333 Munich, Germany
(e-mail: rigoll@tum.de).}
}

\markboth{IEEE TRANSACTIONS ON INTELLIGENT TRANSPORTATION SYSTEMS}%
{Shell \MakeLowercase{\textit{et al.}}: A Sample Article Using IEEEtran.cls for IEEE Journals}


\maketitle

\begin{abstract}
Multi-agent trajectory prediction at signalized intersections is crucial for developing efficient intelligent transportation systems and safe autonomous driving systems. 
Due to the complexity of intersection scenarios and the limitations of single-vehicle perception, the performance of vehicle-centric prediction methods has reached a plateau. 
In this paper, we introduce an Infrastructure-to-Everything (I2X) collaborative prediction scheme. In this scheme, roadside units (RSUs) independently forecast the future trajectories of all vehicles and transmit these predictions unidirectionally to subscribing vehicles. 
Building on this scheme, we propose I2XTraj, a dedicated infrastructure-based trajectory prediction model. I2XTraj leverages real-time traffic signal states, prior maneuver strategy knowledge, and multi-agent interactions to generate accurate, joint multi-modal trajectory prediction. 
First, a continuous signal-informed mechanism is proposed to adaptively process real-time traffic signals to guide trajectory proposal generation under varied intersection configurations. Second, a driving strategy awareness mechanism estimates the joint distribution of maneuver strategies by integrating spatial priors of intersection areas with dynamic vehicle states, enabling coverage of the full set of feasible maneuvers. Third, a spatial-temporal-mode attention network models multi-agent interactions to refine and adjust joint trajectory outputs.
Finally, I2XTraj is evaluated on two real-world datasets of signalized intersections, the V2X-Seq and the SinD drone dataset. In both single-infrastructure and online collaborative scenarios, our model outperforms state-of-the-art methods by over 30\% on V2X-Seq and 15\% on SinD, demonstrating strong generalizability and robustness. The proposed I2X scheme and I2XTraj model offer a flexible solution for enhancing autonomous driving systems with infrastructure-enabled trajectory prediction.

\end{abstract}


\section{Introduction}
\IEEEPARstart{T}{rajectory} prediction is vital in enabling autonomous vehicles to navigate safely and efficiently. Signalized intersections, which account for a disproportionate number of traffic accidents\cite{ekmekci2024effects}, present particularly complex challenges for autonomous systems. Infrastructure-based trajectory prediction systems, enabled by Vehicle-to-Everything (V2X) communication, can provide centralized trajectory prediction services to all vehicles approaching intersections. 
This approach not only enhances autonomous driving capabilities to reduce the occurrence of accidents but also significantly improves V2X communication dependability\cite{zeng2024task}.

\begin{figure}[t]
\flushleft
\includegraphics[width=0.5\textwidth]{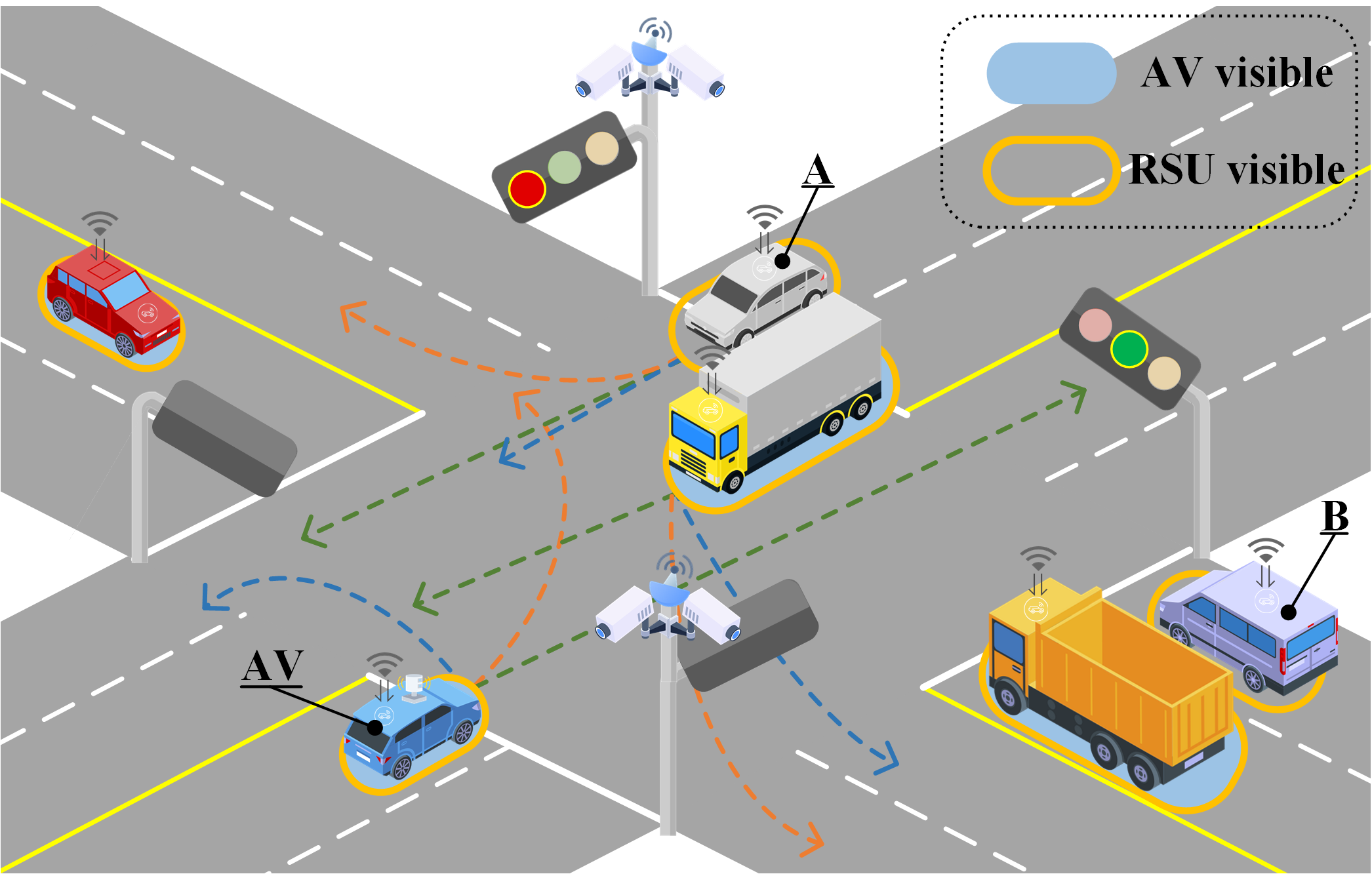}
\caption{Schematic illustration of Infrastructure-to-Everything (I2X) trajectory prediction at a signalized intersection. Autonomous Vehicles (AVs) cannot observe all vehicles. The Roadside Unit (RSU) predicts the joint trajectories of all vehicles at the intersection by leveraging comprehensive vehicle states, real-time traffic signal information, and prior maneuver  patterns. Each set of joint trajectories with the same color represents a possible future scene. These predicted future scenes are transmitted unidirectionally via V2X communication to any vehicle subscribed to the prediction service.}
\label{fig: introduction}
\end{figure}

Many studies propose vehicle-centric trajectory prediction models for highways\cite{liao2024bat}, urban roads\cite{gu2021densetnt}, and unsignalized intersections\cite{rowe2023fjmp}. These models show promising results in various driving scenarios. However, few studies address trajectory prediction at signalized intersections. 
A prototypical signalized intersection scenario is illustrated in Fig. \ref{fig: introduction}, where the roadside unit (RSU) maintains full visibility of all vehicles. The regulation of traffic signal for vehicular flow imposes deterministic constraints on future trajectory prediction, whereas the multi‐branch characteristics of intersection lanes introduce uncertainty to maneuvering strategies available to vehicles. 

Traffic signal information and prior knowledge of maneuvering strategies are not readily obtainable by on-board autonomous driving systems in isolation. With V2X technology, external information from other vehicles and infrastructures has been used to facilitate more accurate and reliable trajectory predictions than vehicle-based systems\cite{ruan2023learning,annunziata2024road}. 
These data sourced from infrastructure may be fused or exchanged at successive stages of an autonomous driving pipeline\cite{10814969}. At the earliest stage, raw image or point‐cloud data are transmitted. Later stages exchange higher-level information, culminating in predicted vehicle trajectories. As processing depth increases, the volume of data requiring transmission between the autonomous vehicle (AV) and roadside units (RSUs) decreases. Consequently, the demands on temporal synchronization and communication latency are progressively reduced\cite{sarkar2023enhancing,wang2024proactive}. 

In prediction‐mediated cooperative schemes, information exchange between vehicle and infrastructure typically proceeds through three stages, including historical trajectories\cite{ruan2023learning}, trajectory features\cite{chen2024conformal,zhang2025co,Co-HTTP}, and future trajectories\cite{annunziata2024road}.  These cooperative schemes typically demand a high degree of coupling between the on-board model and roadside information, constraining the design of autonomous driving systems. To overcome this limitation, we propose an I2X collaborative prediction scheme, which forecasts all intersection vehicle trajectories independently within the infrastructure, then unidirectionally transmits the future trajectories to the vehicles. The roadside-predicted trajectories can serve as input to any type of autonomous driving system, thereby endowing the on-board subsystem with considerable design flexibility.

Within the I2X collaborative prediction scheme, trajectory prediction models deployed on infrastructure play a critical role. On the one hand, the inherent characteristics of such infrastructure information also introduce unique challenges. In particular, the infrastructure is unable to distinguish between autonomous and conventional vehicles traversing an intersection, and may observe multiple autonomous vehicles simultaneously. Consequently, the prediction model must be capable of forecasting the trajectories of all vehicles at the intersection, thereby enabling autonomous vehicles to utilize the provided prediction outputs selectively. 
On the other hand, the supplementary information provided by the infrastructure offers actionable insights for multi-agent trajectory prediction, particularly in scenarios involving a high density of traffic participants. 
This information comprises real-time traffic signal states and prior knowledge of intersection vehicle maneuvering strategies. By modeling the dependence of vehicle trajectories on traffic control signals, the prediction model can reduce uncertainty in forecasts. Moreover, incorporating prior maneuver strategies enables the extraction of latent driving intentions, thereby covering the full set of potential feasible trajectories. 

Therefore, we propose I2XTraj, an infrastructure-based trajectory prediction model for all vehicles at signalized intersections. Our method performs trajectory prediction in a knowledge-driven manner with dynamic graph attention by leveraging real-time traffic information and prior maneuver strategy knowledge from infrastructure devices. Firstly, the \textit{continuous signal-informed mechanism} is designed to represent signal sequences obtained directly from traffic light devices. Different types of intersections have varying rules and deployment of traffic lights. Our nonlinear continuous encoding function adaptively integrates color sequences and control directions to guide the generation of trajectory proposals. Secondly, the \textit{driving strategy awareness mechanism} estimates the joint distribution of maneuver strategies, utilizing prior knowledge of intersection areas and vehicle driving states. Based on the distribution, multi-modal joint trajectory proposals are generated. Thirdly, the \textit{spatial-temporal-mode attention} is employed to model interaction relationships among agents, enabling fine-tuned adjustment of joint predicted trajectories. Finally, we validate our approach on the real-world dataset V2X-Seq and the signalized intersection drone dataset SinD. In single-infrastructure scenarios, our model provides reliable joint multi-agent trajectory predictions for all visible agents. In online collaborative scenarios, our method achieves more accurate single-agent prediction results for the target agent.

The principal contributions of this study are:

(1) The I2X collaborative prediction scheme is proposed, wherein the infrastructure predicts the trajectories of all vehicles at signalized intersections independently and transmits them to subscribing vehicles unidirectionally. 

(2) A dedicated infrastructure knowledge-informed trajectory prediction model in the I2X scheme, I2XTraj, is designed to accurately forecast the joint future trajectories of all vehicles at intersections, leveraging traffic signal states, maneuver strategy knowledge, and interaction information effectively. 

(3) Our method outperforms existing state-of-the-art methods by more than 30\% in the real-world V2I dataset V2XSeq and 15\% in the drone dataset at signalized intersection SinD. The generalizability across different intersections and the robustness are also validated.

The remaining sections of this paper are organized as follows. Section \ref{Related Work} reviews the related work. Section \ref{Problem Formulation} outlines the cooperative scenario and problem formulation. Section \ref{Methodology} details the methodology of our trajectory prediction method. Section \ref{Experiments} thoroughly analyzes the experiment’s results and evaluates the model’s performance. Finally, we summarizes the entire work in Section \ref{Conclusion}.

\section{Related Work}
\label{Related Work}
Most existing trajectory prediction studies focus on general urban environments\cite{ettinger2021large,caesar2020nuscenes} or unsignalized intersections\cite{Argoverse,Argoverse2}, where the contextual information is predominantly obtained through vehicle-based sensors. Vehicle-based methods are constrained by their perceptual capabilities, lacking access to global trajectory information and real-time traffic control signals at intersections, leading to suboptimal performance. Consequently, cooperative trajectory prediction has been recognized as a promising solution to address these challenges\cite{wang2024deepaccident}. In recent years, with the advancement of V2X technologies, several large-scale cooperative datasets that support prediction tasks have been released\cite{yu2023v2x,yan2023int2,zhou2024v2xpnp}.

This section reviews cooperative trajectory prediction at signalized intersections with traffic signals and unique driving maneuvers.

\subsection{Cooperative Trajectory Prediction}
The existing V2X research predominantly focuses on cooperative perception\cite{bai2024survey,xu2023v2v4real,caillot2022survey}, while collaboration approaches leveraging prediction as a nexus have progressively garnered attention among researchers, due to their effective utilization of multi-source information and interaction characteristics. Researchers have established cooperative prediction paradigms at three stages: historical trajectory positions\cite{ruan2023learning}, cooperative features\cite{chen2024conformal,zhang2025co,Co-HTTP}, and layered prediction results\cite{annunziata2024road,cao2024segment}. However, these approaches rely on a highly coupled vehicle–infrastructure cooperative system and ultra-low communication latency, which imposes significant constraints on the development of autonomous driving models. Our method addresses the perceptual limitations by utilizing overhead roadside sensors, simplifying the complex task of joint prediction-communication optimization into a simple public information subscription task for vehicles.

\subsection{Traffic Signal Representations for Trajectory Prediction}
Traffic signal states have been demonstrated to influence driving maneuvers at intersections significantly\cite{paul2022effects}. In early studies, the states have been typically utilized in discrete forms. Discrete indexes have established dependencies for discontinuous vehicle motion behaviors\cite{zhang2022d2}. One-hot encoding has been used to predict vehicle intentions\cite{cao2024segment} and simulate vehicle behaviors\cite{wu2024data}. Dictionaries have embedded discrete states into high-dimensional spaces\cite{wei2024ki}. 
Bayesian networks have also been employed to infer discrete intentions\cite{10133889}.
However, these discrete methods neglect the temporal characteristics of signals. Furthermore, previous methods have been designed for standard orthogonal intersections, lacking generalizability across different intersection types. We propose a continuous signal-informed mechanism adaptively transforms raw traffic light color sequences and control directions into a nonlinear representation that guides trajectory proposal generation under varied intersection configurations.  
\begin{figure*}[t]
\centering
\includegraphics[width=1\textwidth]{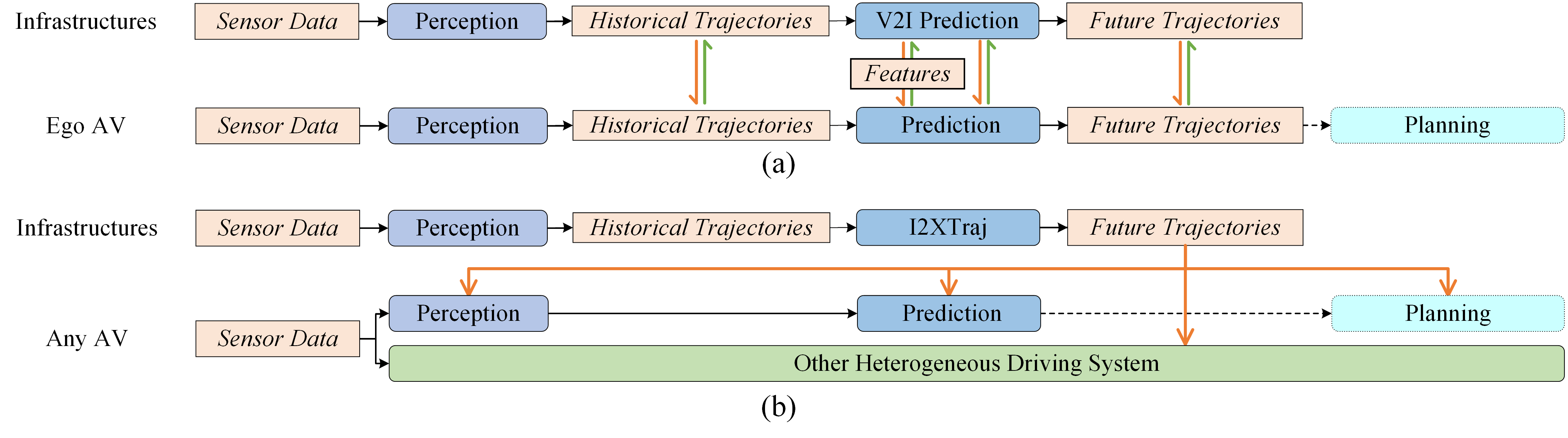}
\caption{A comparison between the Vehicle-Infrastructure Cooperation (VIC) pipeline and our proposed Infrastructure-to-Everything (I2X) cooperative prediction pipeline. (a) The VIC pipeline requires the autonomous vehicle (Ego AV) and infrastructure to be equipped with a unified system, enabling bidirectional communication at one or multiple stages. (b) The I2X cooperative pipeline does not require vehicles to send information to the infrastructure; instead, the infrastructure unidirectionally provides future trajectories to any type of autonomous driving system.}
\label{fig:Scenario}
\end{figure*}

\subsection{Trajectory Prediction with Knowledge Priors}
Effective mining of historical trajectory and traffic flow data can yield essential behavioral priors and patterns\cite{ding2023incorporating}. Such knowledge is typically employed to design secondary tasks, encompassing goal-based\cite{gu2021densetnt}, maneuver-based\cite{ijcai2024p756}, and interaction-based\cite{rowe2023fjmp} trajectory prediction. The incorporation of hard kinematic constraints and soft constraints derived from traffic regulations enhances both the reliability and interpretability of the prediction\cite{10932656}. Infrastructure sensor devices enable the superior collection of maneuver distributions in intersection areas. Leveraging prior knowledge of driving maneuver strategies and intersection topologies, the driving strategy awareness mechanism captures intentions of vehicles, enabling multi-modal prediction to cover all potential maneuvers.

\section{Cooperative Scenario and Problem Formulation}
\label{Problem Formulation}
\subsection{Infrastructure-to-Everything Cooperative Scenario}

In existing collaborative trajectory prediction approaches, Vehicle-Infrastructure Cooperation (VIC) methods are commonly employed, as illustrated in Fig. \ref{fig:Scenario}(a). VIC methodologies can be categorized into two types\cite{yu2023v2x}: online VIC and offline VIC. Online VIC refers to the process where vehicles and infrastructure synchronously conduct real-time data transmission and fusion during collaboration. 
Typical online VIC approaches can be categorized based on the levels at which data fusion is performed. PP-VIC \cite{yu2023v2x} supplies the ego vehicle with infrastructure-side data in a frame-by-frame manner over the historical time horizon. The feature fusion approach \cite{ruan2023learning} leverages motion and interaction features to enable cooperative trajectory feature fusion. The layered approach \cite{annunziata2024road} exploits synergistic collaboration between infrastructure and vehicle layers to enhance prediction performance.
The offline VIC approach first pre-trains models on infrastructure data before deploying them to vehicles for further training. However, both offline and online methods require highly coupled or centralized training between vehicles and infrastructure, which is challenging to implement in practical deployments. This is because real-world scenarios cannot mandate all autonomous vehicles to be equipped with identical cooperative systems or autonomous driving models, nor can they require vehicles to share their real-time data with infrastructure actively. Therefore, we propose the I2X prediction collaboration scheme, illustrated in Fig. \ref{fig:Scenario}(b), which involves deploying specialized prediction models for infrastructure and transmitting the future trajectories to all vehicles. 

Since it is impossible to distinguish whether vehicles are equipped with autonomous driving systems from an infrastructure perspective, it cannot be applied that conventional methods of selecting target agents surrounding the Ego AV on RSUs. This issue has never been addressed in any existing VIC research. Consequently, specialized models deployed on infrastructure need to predict joint future trajectories for all vehicles in the scene and transmit them collectively to subscribing vehicles, allowing these vehicles to select agents of interest independently. Depending on the level of traffic congestion, I2XTraj is required to predict from a few to well over one hundred agents.

This collaborative approach provides considerable flexibility to autonomous vehicles, enabling any type of autonomous driving system to utilize these additional future trajectories from a global perspective. For planning and control systems, these future trajectories can either serve directly as part of the planning input or guide perception or prediction processes. Any other heterogeneous autonomous driving systems can also subscribe to these future trajectories as supplementary information. 

\subsection{Problem Definition}

\begin{figure*}[t]
\centering
\includegraphics[width=1\textwidth]{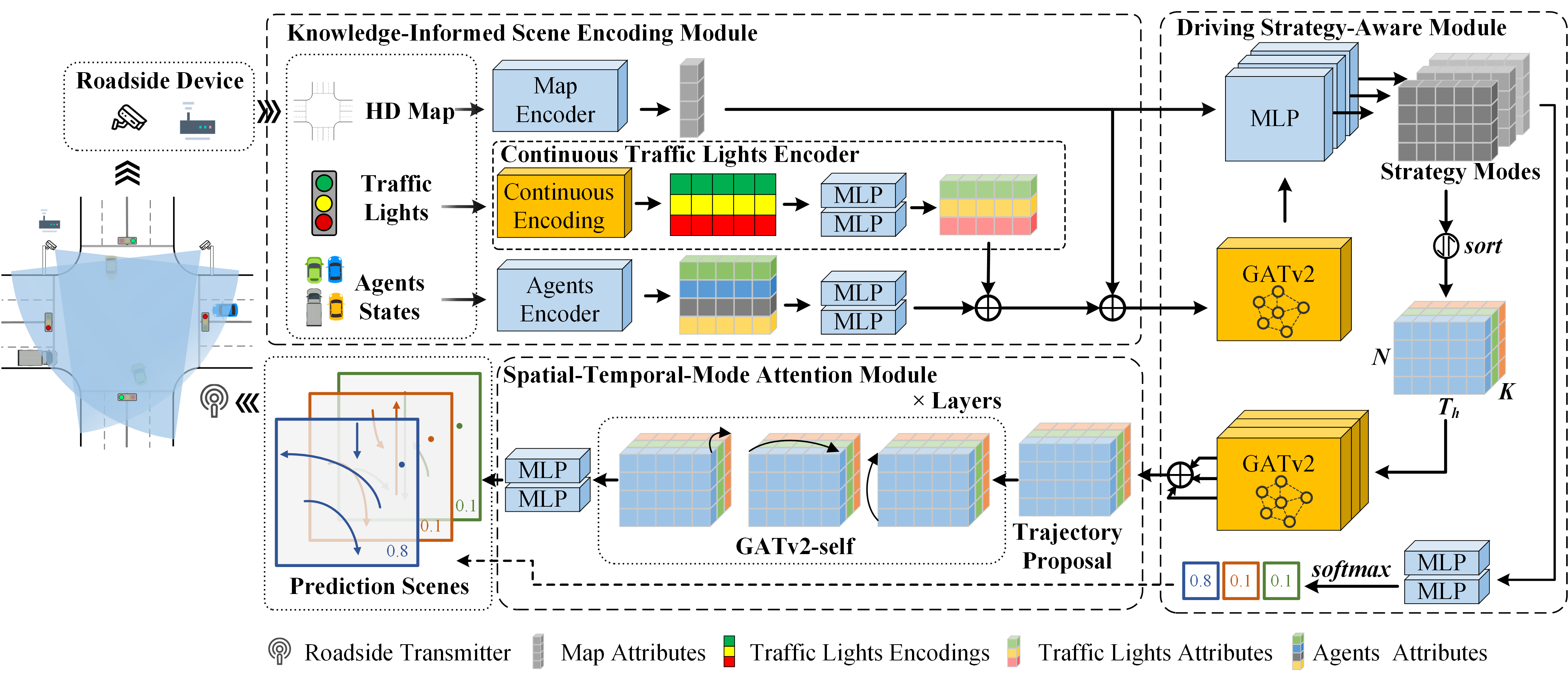}
\caption{The overall framework of our I2XTraj. Our architecture is an infrastructure-based method, which comprises three parts: (a) Knowledge-Informed Scene Encoding Module embeds agents' historical states with traffic signal and map knowledge. (b) Driving Strategy-Aware Module generates strategy modes based on topological features and maneuver strategies distributions to trajectory proposals. (c) Spatial-Temporal-Mode Attention Module spans the three dimensions to generate predicted scene trajectories.  }
\label{fig:framework}
\end{figure*}

The objective of multi-agent trajectory prediction is to forecast the future trajectories of all agents with historical states. The model can directly extract the state features of agents $\mathrm{H}$ from the infrastructure within historical time steps $T_{h}$. Given that specific past states can lead to multiple possible futures, the maneuver of joint agents $N$ prediction is to predict trajectories for any agent $n$ under each possible scene ${S}^{k}$:
\begin{align}
    {Y} = \left\{ {{S}^{k}} \right\} = \{{F}^{0,k},{F}^{1,k},\cdots,{F}^{n,k}\}_{k \in [0,K-1]},
\end{align}%
where $K$ denotes scene modes. Additionally, traffic light signals ${L}$ and high-definition (HD) maps ${M}$ are incorporated into the model as prior knowledge ${I}$. Formally, a joint future trajectory distribution among agents is represented as:
 \begin{align}
    P({S}^{k}|{H},{I}) = \prod_{n=0}^{N-1} P({F}^{n,k}|{H},{L},{M}),
\end{align}%
where ${F}^{n,k} = \left\{ {p_{x}^{t_c:t_c+T_{f}},p_{y}^{t_c:t_c+T_{f}}} \right\}$ represents predicted future positions over the next $T_{f}$ time steps at current time step $t_c$.

\section{Methodology}
\label{Methodology}

The proposed I2XTraj framework comprises three integral components, as illustrated in Fig. \ref{fig:framework}: the Knowledge-Informed Scene Encoding Module, the Driving Strategy-Aware Module, and the Spatial-Temporal-Mode Attention Module. Each component is specifically engineered to address the unique challenges of infrastructure-based trajectory prediction at signalized intersections.

\subsection{Knowledge-Informed Scene Encoding Module}

Infrastructure-based trajectory prediction primarily requires encoding of intersection scenarios. Recent studies have demonstrated the remarkable effectiveness of Graph Neural Networks (GNNs) and relative spatiotemporal position encoding\cite{tang2024hpnet,zhou2023query}. In the knowledge-informed scene encoding module, agent states, traffic light states, and HD maps are encoded as node attributes, while their relative spatio-temporal relationships are represented as edge attributes. This encoding method enables the framework to capture both the individual characteristics of scene elements and their intricate interconnections within the dynamic intersection environment.  

\paragraph{Agent Encoder} The Agent Encoder module systematically encodes the agent states, including the spatial positions, motion states, geometric dimensions, and semantic attributes of agents within the scene. Specifically, a two-layer MLP is employed to embed the state of each agent at time step $ t\in[t_c-T_h,t_c]$ into the agent attribute:
\begin{align}
    {{A}_A^{t,n}} = \text{MLP}(p_x^{t,n}, p_y^{t,n}, \theta^{t,n}, \rho^{t,n}, \varphi^{t,n}, l^{t,n}, w^{t,n} ,c^{t,n}),
\end{align}%
where \( \left( p_{x}^{t, n}, p_{y}^{t, n} \right) \) is the location, \( \theta^{t, n} \) is the orientation, \( \left( \rho^{t,n}, \varphi^{t,n} \right) \) represent the velocity in polar coordinates defined with the current position as the origin and direction as the positive axis,\( \left( l^{t,n}, w^{t,n} \right) \) denote length and width, and \( c^{t, n} \) is the type attribute. 

\begin{figure}[t]
\centering
\includegraphics[width=0.5\textwidth]{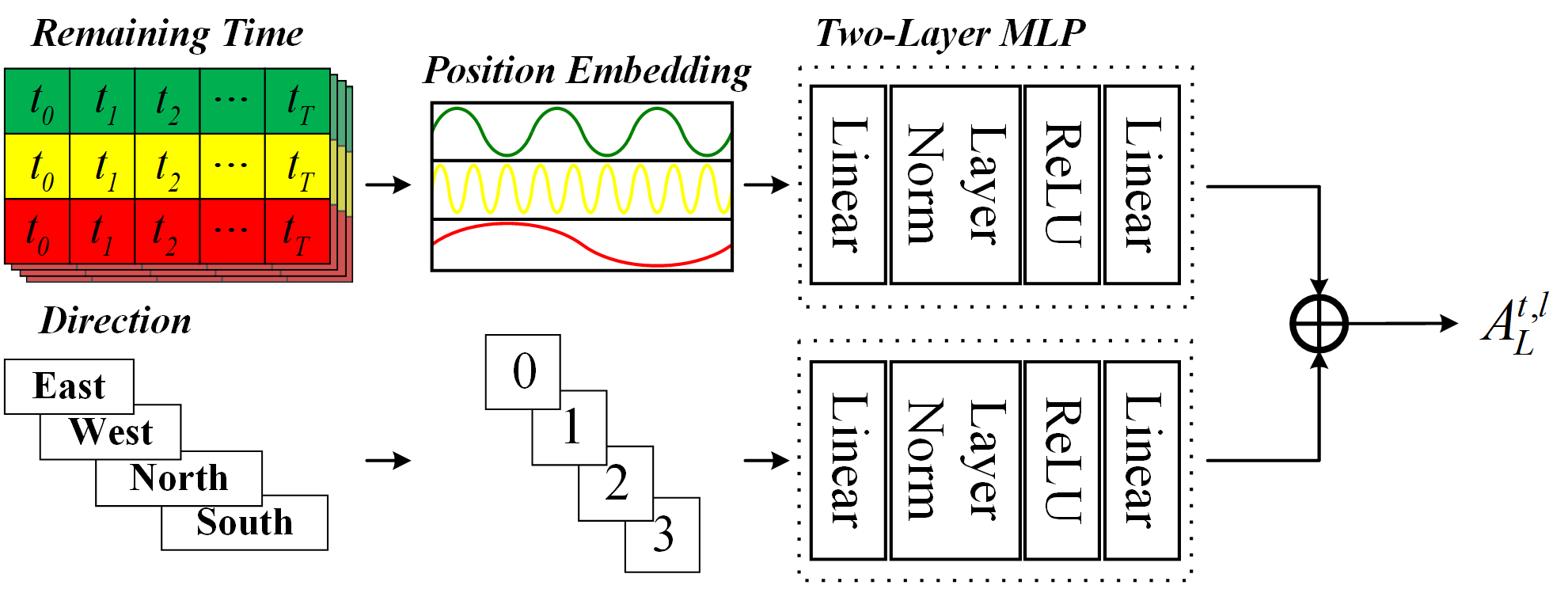}
\caption{Schematic illustration of the continuous signal-informed mechanism.}
\label{fig: tl encoder}
\end{figure}

\paragraph{Continuous Traffic Light Encoder}I2XTraj deployed directly on infrastructure can conveniently utilize traffic light information, including its continuous digital signals, control directions, and traffic light locations. As shown in Fig. \ref{fig: tl encoder}, we encode the digital signals using a nonlinear continuous function similar to positional embedding:
\begin{equation}
\begin{aligned}
{PE}^{t,\ell} &= \sin(\frac{t_{remain}^{t, d, \ell}}{T_{\Omega} \cdot \left(\frac{1}{3}\right)^{d}}),
\end{aligned}
\end{equation}
where $t_{remain}^{t, d, \ell}$ is the remaining time of each color signal of a single traffic light $\ell$, $T_{\Omega}$ is the maximum cycle time of the traffic light sequence, and $d \in\left\{0,1,2\right\}$ indicates the color of signals, typically following the order of red, green, and yellow.

The advantage of this approach is its adaptation to the different traffic control patterns at various intersections. Furthermore, the nonlinear function causes a sharp increase in the rate of change as the remaining time approaches zero, making the model more sensitive to the instantaneous transition of traffic light signals. Generally, the maximum cycle time $T_{\Omega}$ ensures the monotonicity of the values. For intersections with unknown traffic light patterns or controlled by intelligent transportation systems, a sufficiently large value has been proven to be effective.

The traffic flow direction controlled by the traffic light and the encoded signal are embedded using a two-layer MLP and then combined into traffic light attributes:
\begin{align}
    {{A}_L^{t,\ell}} = \text{MLP}({PE}^{t,\ell})+ \text{MLP}({d}_L),
\end{align}%
where $d_L$ is the direction of traffic flow controlled by the traffic light.
The composite feature representation encapsulates the traffic signal-governed intersection scenarios. This integrated representation captures both the instantaneous state of the intersection under traffic control and maintains the temporal context necessary for understanding the evolution of traffic patterns.

 \paragraph{Map Encoder}Beyond the typical spatial features of lane segments such as position, length, and adjacency relationships, intersection maps contain rich semantic attributes, including interior intersection indicators, turning directions, and signal control states. A two-layer MLP is employed to encode the map attributes: 
\begin{align}
    {{A}_M} = \text{MLP}({l}_{M}, {c}_{M}),
\end{align}%
where $l_{M}$ represents the length of the centerlines, and ${c}_{M}$ denotes the attribute features of the lane segment. The attribute features include lane center position, lane heading, is intersection, turning sign, is controlled by traffic lights, centerline positions, and centerline heading. All of these features come from the HD map automatically. The positions and coordinates of the centerlines are embedded as edge features. LaneGCN\cite{liang2020learning} is employed to extract the topological relationships between lanes, utilizing graph self-attention to capture the interaction features.  

\paragraph{Graph Edge Encoder}Similar to HPNet\cite{tang2024hpnet}, we encode the relative spatio-temporal relationships between nodes as edge features in local polar coordinate systems. A two-layer MLP is employed to encode edge features:
\begin{align}
    {E}_{e} = \text{MLP}({d}_{e}, {\phi}_{e}, {\psi}_{e}, {\delta}_{e}),
\end{align}%
where ${d}_{e}$, ${\phi}_{e}$, ${\psi}_{e}$ and ${\delta}_{e}$ represent spatial distance, the edge orientation in the reference frame, relative node orientation and temporal difference, respectively. 

\begin{figure}[t]
\centering
\includegraphics[width=0.4\textwidth]{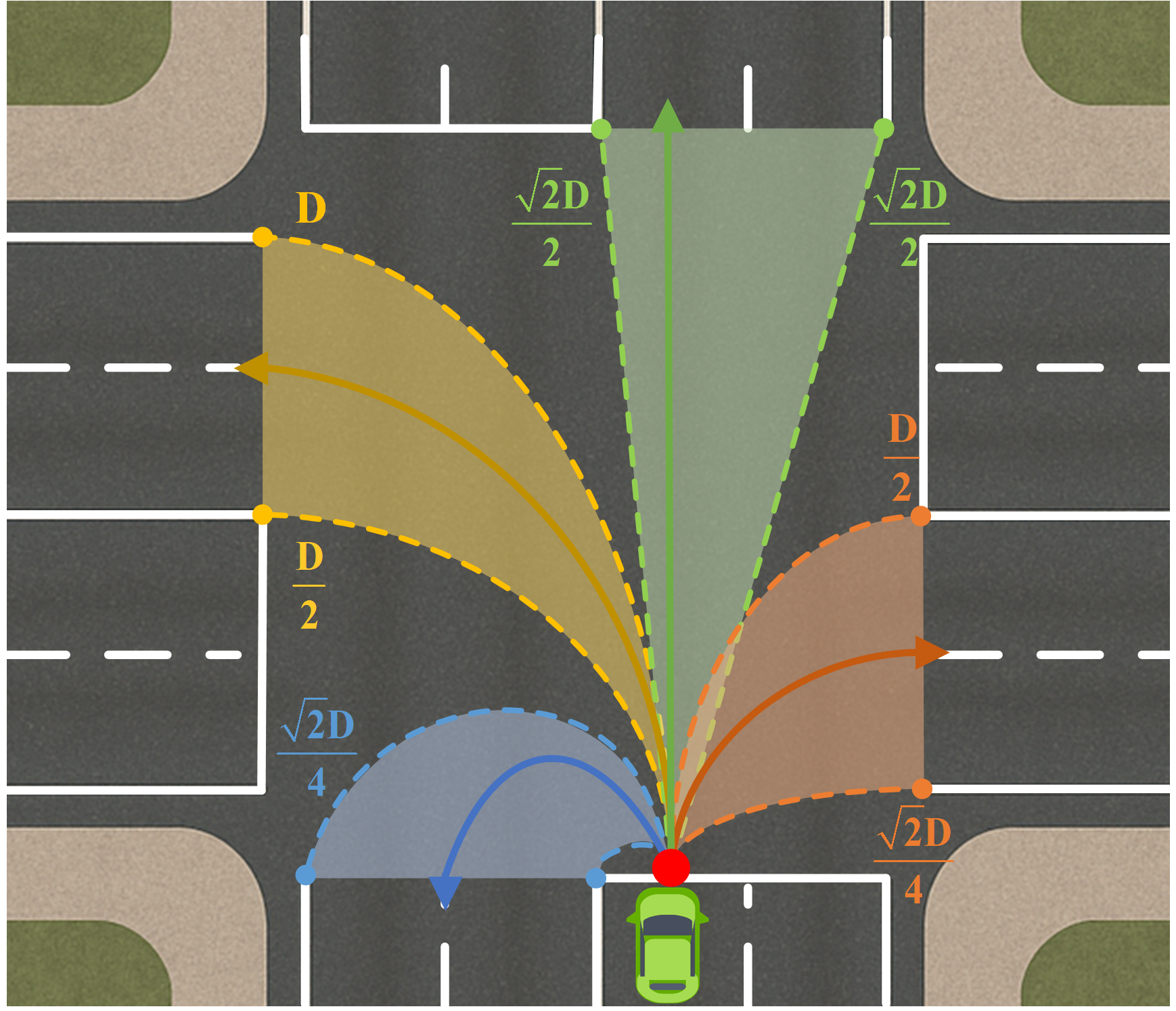}
\caption{Schematic illustration of the driving awareness mechanism. The red circular area indicates the range of the stop maneuvers. The blue area represents the range of U-turn maneuvers. The yellow area corresponds to the range of left-turning maneuvers, while the green area denotes the range of maneuvers moving straight. The orange area indicates the range of right-turning maneuvers.}
\label{fig: Strategy}
\end{figure}

\subsection{Driving Strategy-Aware Module}

The trajectory data distributions derived from infrastructure-captured and vehicle-captured sources exhibit substantial disparities. Firstly, the field of view of roadside sensors exclusively encompasses the traffic conditions within the intersection area. Secondly, due to the complex nature of intersection environments, vehicles naturally tend to reduce their speed upon entering the intersection. Thirdly, vehicle maneuvers at intersections are constrained to a finite set of options, typically comprising stopping, through movement, left/right turns, and U-turns.

The driving awareness mechanism generates potential maneuver intervals based on the prior intersection diameter $D$ knowledge and simple geometric operations.
The stop maneuver corresponds to the interval $[0,1)$. The go straight is represented by the interval $[\sqrt{2}D,+\infty)$. Right turns are associated with the interval $[\frac{\sqrt{2}D}{2},\frac{D}{2})$. Left turns are defined within the interval $[\frac{D}{2}, D)$. U-turns correspond to the interval $[1,\frac{\sqrt{2}D}{2})$. These interval assignments are illustrated in Fig. \ref{fig: Strategy}.
The strategies for multi-modal intersection maneuver are derived from latent driving intentions, thus the driving awareness mechanism maps these maneuver intervals into distinct modes. Corresponding to $K$ modes, the $K$ maneuver intervals  $\{[0,1),[1,\frac{D}{K-2}), $
$[\frac{D}{K-2},\frac{2D}{K-2}),\cdots,[\frac{(k-3)D}{K-2},D),[D,+\infty)\}$ encompasses various intersection maneuvers. 

\begin{figure}[t]
\flushleft
\includegraphics[width=0.5\textwidth]{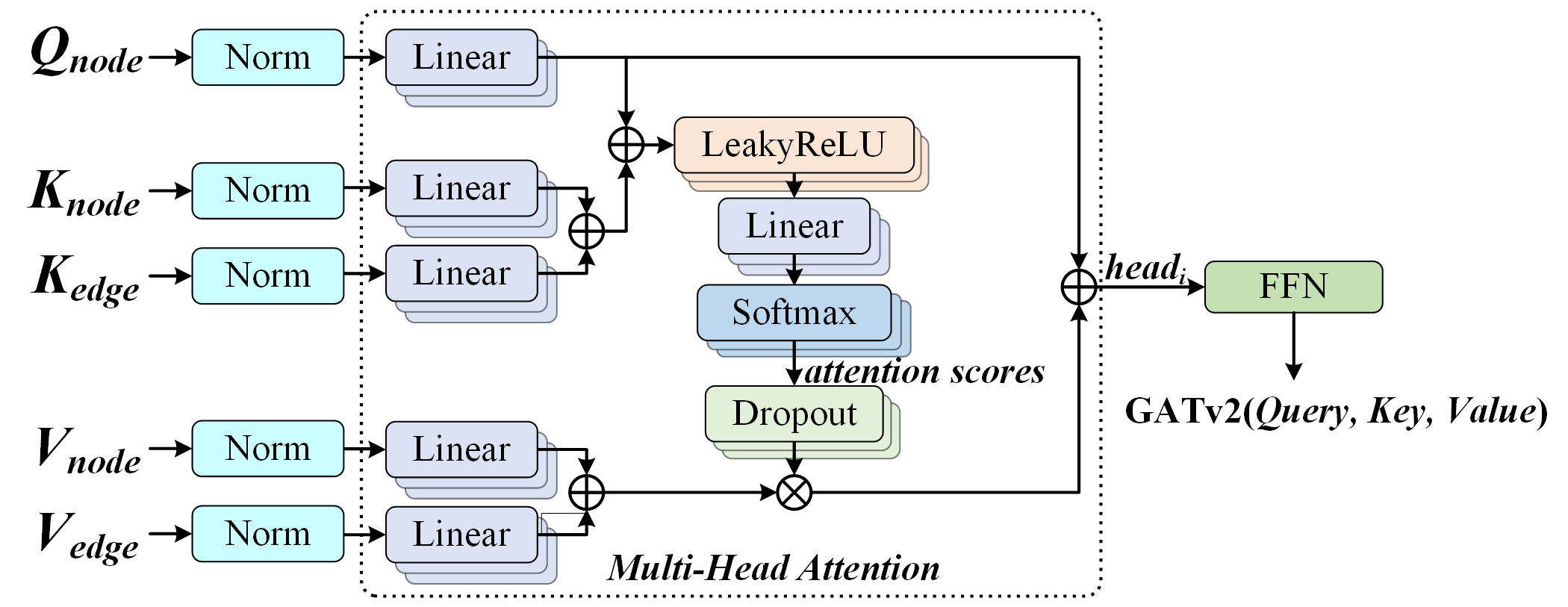}
\caption{Schematic illustration of dynamic graph attention. FFN donates Feed Forward Network layer.}
\label{fig: gatv2}
\end{figure}

The strategy modes are predicted for all agents at the intersection in parallel. In essence, this involves mapping spatio-temporal attributes onto each maneuver interval by dynamic graph attention. The driving awareness decoder component comprises a self-attention module and $K$ independent two-layer MLP decoders. Similar to the object query concept in DETR\cite{carion2020end}, each independent decoder generates adaptive a trajectory anchor query $Q^{t,n}$ corresponding to each mode query $q_K^{t,n,k}$:
\begin{align}
    Q^{t,n} = \text{GATv2}({A_A^{t,n}}, [{A_A^{t,n}},E_e], [{A_A^{t,n}},E_e]),
\end{align}%
\begin{align}
    q_K^{t,n,k} =\text{MLP}(Q^{t,n}),
\end{align}%
where $\text{GATv2} (Query, Key, Value)$ denotes the dynamic graph attention\cite{brody2022how}, as Fig. \ref{fig: gatv2}. These queries are processed through an MLP output layer to produce the probability scores $P_k$ of strategy modes. Specifically, $P_k$ is assumed the probabilities of maneuvers falling within intervals. 
\begin{align}
    P_k =softmax(\text{MLP}(q_K^{t,n,k})).
\end{align}%
The mode queries are sorted based on the scores and stacked into mode anchor attributes ${A_K^{t,n,k}}$.

The mode anchor attributes subsequently interact with agents, traffic signals, and map features through cross-attention modules. During the dynamic process, the trajectory distribution diverges when the agent enters the intersection, and gradually converges as it exits the intersection area. To capture this trend, agent proposal query $q_A^{t,n,k}$,  signal proposal query $q_L^{t,n,k}$ and map proposal query $q_M^{t,n,k}$ are generated:
\begin{equation}
\begin{aligned}
    q_A^{t,n,k} = \text{GATv2}(&{A_K^{t,n,k}}, [{A_A^{t_c-T_h:t,n}},E_e], \\
   & [{A_A^{t_c-T_h:t,n}},E_e]),
\end{aligned}
\end{equation}
\begin{equation}
\begin{aligned}
    q_L^{t,n,k} = \text{GATv2}(&{A_K^{t,n,k}}, [{A_L^{t_c-T_h:t,\ell}},E_e], \\
   & [{A_L^{t_c-T_h:t,\ell}},E_e]),
\end{aligned}
\end{equation}
\begin{equation}
\begin{aligned}
    q_M^{t,n,k} = \text{GATv2}({A_K^{t,n,k}}, [A_M,E_e],
    [A_M,E_e]),
\end{aligned}
\end{equation}
These queries are summed to generate the predicted trajectory proposal attribute  $Q_{P}^{t,n,k}$:
\begin{align}
    {Q_{P}^{t,n,k}} = q_A^{t,n,k} + q_L^{t,n,k} + q_M^{t,n,k}.
\end{align}%

\subsection{Spatial-Temporal-Mode Attention Module}
The Spatial-Temporal-Mode Attention Module enables predictions to interact with any agents, any historical timestamps, and any modes. These interactions can be either direct or multi-hop. For the spatiality, an agent $A$ performs cross-attention with each mode at every timestamp of every neighboring agent in the scene:
\begin{equation}
\begin{aligned}
    {Q_{A}^{t,n,k}} = \text{GATv2}({Q_P^{t,n,k}}, [{Q_P^{t,n_{nbr},k}},E_e], [{Q_P^{t,n_{nbr},k}},E_e]),
\end{aligned}
\end{equation}
where $n_{nbr}$ denotes all neighboring agents. For the temporal, each timestamp $T$ performs self-attention with each mode of every agent at every historical timestamp:
\begin{equation}
\begin{aligned}
    {Q_{T}^{t,n,k}} = \text{GATv2}(&{Q_A^{t,n,k}}, [{Q_A^{t-t_c+T_h:t,n,k}},E_e],\\
    &[{Q_A^{t-t_c+T_h:t,n,k}},E_e]).
\end{aligned}
\end{equation}
For the mode, each mode $K$ performs self-attention with each mode of every agent at every historical timestamp:
\begin{equation}
\begin{aligned}
    {Q_{K}^{t,n,k}} = \text{GATv2}({Q_T^{t,n,k}}, [{Q_T^{t,n,0:K-1}},E_e], [{Q_T^{t,n,0:K-1}},E_e]).
\end{aligned}
\end{equation}

Multiple iterations enable the learning of broader interaction features across all dimensions while enhancing trajectory prediction accuracy\cite{zhou2023query}.

Finally, a two-layer MLP is employed for trajectory decoding to generate scene prediction trajectories:
\begin{align}
    S^{k}=\text{MLP}({Q_{K}^{t,n,k}}).
\end{align}%

\subsection{Training Objectives}
While the adaptive anchor-based decoding mechanism demonstrates notable advantages in enhancing predictive flexibility, it inherently introduces rationality constraints, with boundary violations emerging as the most substantial limitation. To mitigate this issue, we implement an off-map loss function. The vector map representation transforms a rasterized drivable raster map $M_r$. Following this conversion, we query the spatial coordinates to generate off-map matrices for both the ground truth $O_{gt}^{t,n}=M_r({g_{x}^{t,n}},{g_{y}^{t,n}})$, and predicted trajectories $O_{p}^{t,n,k}=M_r({p_{x}^{t,n,k}},{p_{y}^{t,n,k}})$.
To account for and exclude off-map occurrences attributable to both the vehicle's intrinsic kinematic constraints and potential map imperfections, the off-map loss is computed as follows:
\begin{align}
    \mathcal L_{off-map}=\sum_{t=t_c}^{t_c+T_f}\frac{\sum_{k=0}^{K-1}\sum_{n=0}^{N-1}[O_{p}^{t,n,k}\wedge O_{gt}^{t,n}]}{K \sum_{n=0}^{N-1}O_{gt}^{t,n}}.
\end{align}%
This loss metric is designed to differentiate between legitimate trajectory deviations and those arising from prediction errors.

Following the existing works \cite{zhao2021tnt,rowe2023fjmp}, the joint regression loss is computed utilizing the Huber loss function:
\begin{align}
    \mathcal L_{reg}=\mathcal L_{huber}(S^k,S_{gt}),
\end{align}%
where $S_{gt}$ denotes the ground truth trajectories.

The loss for strategy probabilities $P_{k}$ is computed using cross-entropy:
\begin{align}
    \mathcal L_{cls}=\mathcal L_{CE}(P_{k}, P_{gt}),
\end{align}%
where $P_{gt}$ denotes the ground truth strategies set.

The entire model is optimized using a comprehensive loss function:
\begin{align}
    \mathcal L=(1+\mathcal L_{off-map}) \mathcal L_{reg} + \mathcal L_{cls}.
\end{align}%
In addition, in marginal prediction tasks, the joint regression loss degenerates into the marginal regression loss.

\section{Experiments}
\label{Experiments}
\subsection{Experimental Setup}
\paragraph{Dataset} 
To rigorously evaluate the performance of our model, we select two complementary datasets that explicitly incorporate traffic signal data:
the real-world V2I dataset V2X-Seq\cite{yu2023v2x} and the drone dataset at signalized intersection SinD\cite{xu2022drone}. The V2X-Seq (Single-Infrastructure, SI)  set contains comprising 55,197 pure infrastructure-based scenarios. The V2X-Seq (Cooperation, C) set contains 51,146 V2I scenarios. Both sets feature trajectories of 10-second duration sampled at 10Hz, covering 672 hours of data from 28 intersections. The prediction task involves observing 5 seconds to forecast the subsequent 5 seconds. The SinD dataset contains 7 hours of continuous trajectory data sampled at 10Hz from a signalized intersection in Tianjin, China. It includes trajectories and semantic annotations for over 13,000 traffic participants, along with traffic signal states. The prediction task involves observing 1.2 seconds (12 frames) to predict the subsequent 1.2 seconds (12 frames) and 1.8 seconds (18 frames). 

\paragraph{Metrics} 
The prediction performance is evaluated using a comprehensive set of trajectory metrics, including minADE, minFDE, MR, minJointADE, minJointFDE, and minJointMR. The minADE measures the average L2 distance between predicted and ground truth trajectory points, while minFDE examines the L2 distance at trajectory endpoints. The MR metric calculates the ratio of cases where minFDE exceeds 2 meters. The minJointADE evaluates the average L2 distance between predicted and ground truth trajectories across all agents, while minJointFDE focuses on the L2 distance for all agents at the final timestep. The minJointMR metric calculates the ratio of cases where minJointFDE exceeds 2 meters.  The number of modes $K$ is selected as 6.

\paragraph{Baselines}
 I2XTraj is compared to twelve state-of-the-art (SOTA) trajectory prediction models: TNT\cite{zhao2021tnt}, DenseTNT\cite{gu2021densetnt}, HiVT\cite{zhou2022hivt}, V2INet\cite{chen2024conformal}, V2X-Graph\cite{ruan2023learning}, AIoT\cite{annunziata2024road}, HPNet\cite{tang2024hpnet}, Co-HTTP\cite{Co-HTTP}, Co-MTP\cite{zhang2025co}, HTSI\cite{10947614}  for the V2X-Seq dataset.
 S-GAN\cite{gupta2018social}, S-LSTM\cite{alahi2016social}, Trajetron++\cite{salzmann2020trajectron++}, FJMP\cite{rowe2023fjmp}, KI-GAN\cite{wei2024ki} for SinD dataset.  FF denotes feature fusion method\cite{ruan2023learning}.
 As I2XTraj represents the first work to perform multi-agent prediction tasks on V2X-Seq, we reproduce and compare against HPNet, the SOTA method on the unsignalized intersection dataset INTERACTION, as our baseline. Additionally, HPNet is also adapted to the signalized intersection drone dataset SinD to establish a comparative baseline.  

\paragraph{Implementation Details}
In our implementation, each dynamic graph attention consists of four layers of multi-head attention. Our model is trained on a single A800 GPU for 32 epochs, using the AdamW\cite{loshchilov2018decoupled} optimizer with a batch size of 4, using an initial learning rate of  $5 \times 10^{-4}$, setting 30 meters radius for all local areas. The parameter size of I2XTraj is 3.18 M.  I2XTraj requires a total processing time of 112.23 milliseconds. 

\subsection{Comparison with State-of-the-art}
I2XTraj is compared against the strongest baselines in the V2X-Seq trajectory forecasting benchmarks. Our trajectory prediction framework is developed and evaluated using the V2X-Seq (SI). Given that only one previous study\cite{annunziata2024road} has provided a reference using the single-infrastructure dataset V2X-Seq (SI) for training, we further train and validate our model on the more extensively researched collaborative dataset V2X-Seq (C).

\paragraph{Single-Agent Scenario}
I2XTraj is compared with single-agent SOTA baselines on V2X-Seq. Our framework consistently achieves superior performance across all metrics, as shown in Table \ref{tab:table1}. 

For collaborative scenarios, the PP-VIC method is employed for vehicle-infrastructure data fusion\cite{yu2023v2x}. PP-VIC provides trajectory data from the infrastructure to the ego vehicle in a frame-by-frame manner within the historical range. Our I2XTraj framework demonstrates significant performance improvements compared to the graph-based feature fusion method V2X-Graph, achieving a 26.25\% reduction in minFDE and a 48\% decrease in MR.  Compared to the layer-wise fusion method AIoT, our framework demonstrates a substantial improvement of 18.94\% in minADE. Although PP-VIC is not currently the best performance collaborative method \cite{zhang2025co}, it retains complete infrastructure-captured trajectory and traffic light data, which highlights our model's significant advantages at signalized intersections.

For single-infrastructure scenarios, our infrastructure-based I2XTraj framework underscores more pronounced advantages compared to general methods. Specifically, I2XTraj achieves a 40.16\% reduction in minADE and a 47.35\% reduction in minFDE compared to HiVT. When compared to AIoT, the MR decreases by 47.73\%. The notably higher improvement in MR metrics emphasizes the significance of incorporating off-map loss. Although the overall performance of all methods, including I2XTraj, shows some degradation compared to cooperative scenarios, this may be attributed to the strong correlation between the target agent selection and the ego vehicle in the dataset, where vehicle-captured data provides more accurate trajectories and velocities. However, I2XTraj exhibits substantially smaller average degradation compared to general models, and even a slight improvement in minADE, indicating effective utilization of infrastructure data characteristics. 

\paragraph{Multi-Agent Scenario}
I2XTraj is compared with multi-agent baselines on V2X-Seq (SI) and SinD. Our framework consistently achieves superior performance in most cases, as shown in Table \ref{tab:table2} and Table \ref{tab:table3}.

On the V2X-Seq (SI) dataset, our I2XTraj framework demonstrates superior performance compared to HPNet, the SOTA trajectory prediction method developed for unsignalized intersections. Specifically, our framework achieves significant improvements across all metrics: a 23.53\% reduction in minJointADE, a 30.8\% reduction in minJointFDE, and a 19.05\% reduction in minJointMR. 

On the SinD dataset, our I2XTraj framework also performances improved prediction accuracy, with particularly notable performance in long-term prediction, achieving reductions of 9.09\% and 19.23\% in minADE and minFDE, respectively. While the short-term prediction shows marginal limitations, these results sufficiently support the generalizability of I2XTraj  across heterogeneous data sources.

\begin{table}
    \small
    \centering
    \caption{Performance comparison of single-agent on the V2X-Seq dataset\label{tab:table1}}
    \begin{tabular}{llccc}
        \toprule
        Method    & Dataset/Fusion   & minADE  & minFDE & MR \\
        \midrule
        TNT       & C/PP-VIC & 4.36      & 9.23   &  0.62  \\
        DenseTNT  & C/PP-VIC & 1.84      & 2.56   &  0.28  \\
        HTSI      & C/PP-VIC & 1.45      & 2.67   &  0.43  \\
        HiVT      & C/PP-VIC & 1.27      & 2.36   &  0.30  \\
        V2INet    & C/PP-VIC & 1.19      & 1.98   &  0.27  \\
        V2X-Graph & C/PP-VIC & 1.12      & 1.98   &  0.30  \\
        V2X-Graph & C/FF     & 1.05      & 1.79   &  0.25  \\
        AIoT      & C/Layered     & 0.95      & 1.87   &  0.27  \\
        Co-HTTP   & C/PP-VIC      & 0.91      & 1.46   &  0.20  \\
        Co-MTP    & C/PP-VIC      & \underline{0.85}      & \underline{1.33}   &  \underline{0.18}  \\
        \textbf{I2XTraj}   & C/PP-VIC & \textbf{0.80}      & \textbf{1.27}    &  \textbf{0.12}   \\
        \midrule
        TNT     & SI/-    & 4.93      & 9.45     &  0.65 \\
        AIoT    & SI/-   & 1.36      & 2.96     &  \underline{0.44} \\
        HiVT    & SI/-   & \underline{1.27}    & \underline{2.83}  &  0.47 \\
        \textbf{I2XTraj} & SI/-     & \textbf{0.76}      & \textbf{1.49}  & \textbf{0.23}   \\
        \bottomrule
    \end{tabular}
\end{table}

\begin{table}
    \small
    \caption{Performance comparison of multi-agent on the V2X-Seq (SI) dataset\label{tab:table2}}
    \centering
    \begin{tabular}{lcccc}
        \toprule
        Method   & minJointADE  & minJointFDE & minJointMR \\
        \midrule
        HPNet         & 0.68      & 1.62  & 0.21     \\
        \textbf{I2XTraj}      & \textbf{0.52}   & \textbf{1.12}  & \textbf{0.17}     \\
        \bottomrule
    \end{tabular}
\end{table}

\begin{table}
    \caption{Performance comparison of multi-agent on the SinD dataset \label{tab:table3}}
    \centering
    \begin{tabular}{lcccc}
        \toprule
        \multirow{2}{*}{Method} & \multicolumn{2}{c}{12-12} & \multicolumn{2}{c}{12-18} \\
        \cmidrule(lr){2-3} \cmidrule(lr){4-5}
         & minADE & minFDE & minADE & minFDE \\
        \midrule
        S-GAN & 1.32 & 2.46 & 1.53 & 2.95 \\
        S-LSTM & 0.87 & 1.60 & 0.96 & 1.78 \\
        Trajetron++ & 0.37 & 0.93 & 0.70 & 1.91 \\
        FJMP & 0.27 & 0.68 & 0.41 & 1.13 \\
        KI-GAN & \textbf{0.05} & \textbf{0.12} & 0.11 & 0.26 \\
        HPNet & \underline{0.09} & 0.21 & \underline{0.11} & \underline{0.25} \\
        \textbf{I2XTraj} & \underline{0.09} & \textbf{0.12} & \textbf{0.10} & \textbf{0.21} \\
        \bottomrule
    \end{tabular}
\end{table}

\subsection{Ablation Study}
To further demonstrate the effectiveness and outcomes of the continuous signal-informed and driving strategy awareness mechanism, we conducted ablation studies on the V2X-Seq (SI) dataset. The quantitative results of our ablation studies are presented in Table \ref{tab: Ablation Table}.

Comparing the complete framework against that without the continuous traffic light encoder, we validate the effectiveness of the continuous signal-informed  (CSI) mechanism. The continuous traffic light encoding enables the model to accurately anticipate signal phase transitions, thereby precisely determining vehicles' instantaneous dynamics during acceleration and deceleration events. Additionally, the relative positioning between traffic lights and vehicles helps the model predict specific vehicle stopping locations. The incorporation of traffic light information improved prediction accuracy by 2.61$\sim$7.44\%, with particularly pronounced enhancements in final point precision. 

The Driving Strategy-Aware (DSA) mechanism and map information (MAP) prove crucial for the model's scene comprehension and prediction capabilities. The improvements attributed to DSA validate the effectiveness of joint trajectory prediction based on combined the maneuver intents and maneuvers relationship. The knowledge-driven driving strategy prediction, followed by trajectory mode generation based on strategy confidence scores, provides causal guidance for the prediction. The enhancement brought by MAP demonstrates the significant influence of map topology on vehicle trajectories at intersections. The integration of map topology and maneuver priors enables more accurate vehicle trajectory prediction, yielding improvements of up to 8.00\%.

\begin{table}
    \small
    \caption{Ablation studies of multi/single-agent for core components in V2X-Seq (SI) dataset}
    \centering
    \begin{tabular}{cccccc}
        \toprule
        CSI & MAP & DSA & minADE & minFDE & MR \\
        \midrule
         & \checkmark & \checkmark & 0.55/0.78 & 1.21/1.53  & 0.18/0.24   \\
        \checkmark &  & \checkmark & 0.53/0.79 & 1.14/1.56 & 0.17/0.28  \\
        \checkmark & \checkmark &  & 0.53/0.77 & 1.16/1.52 & 0.18/0.24   \\
        \checkmark & \checkmark & \checkmark & \textbf{0.52}/\textbf{0.76} & \textbf{1.12}/\textbf{1.49} & \textbf{0.17}/\textbf{0.23}   \\
        \bottomrule
    \end{tabular}
    \label{tab: Ablation Table}
\end{table}

\begin{figure}[t]
\flushleft
\includegraphics[width=0.5\textwidth]{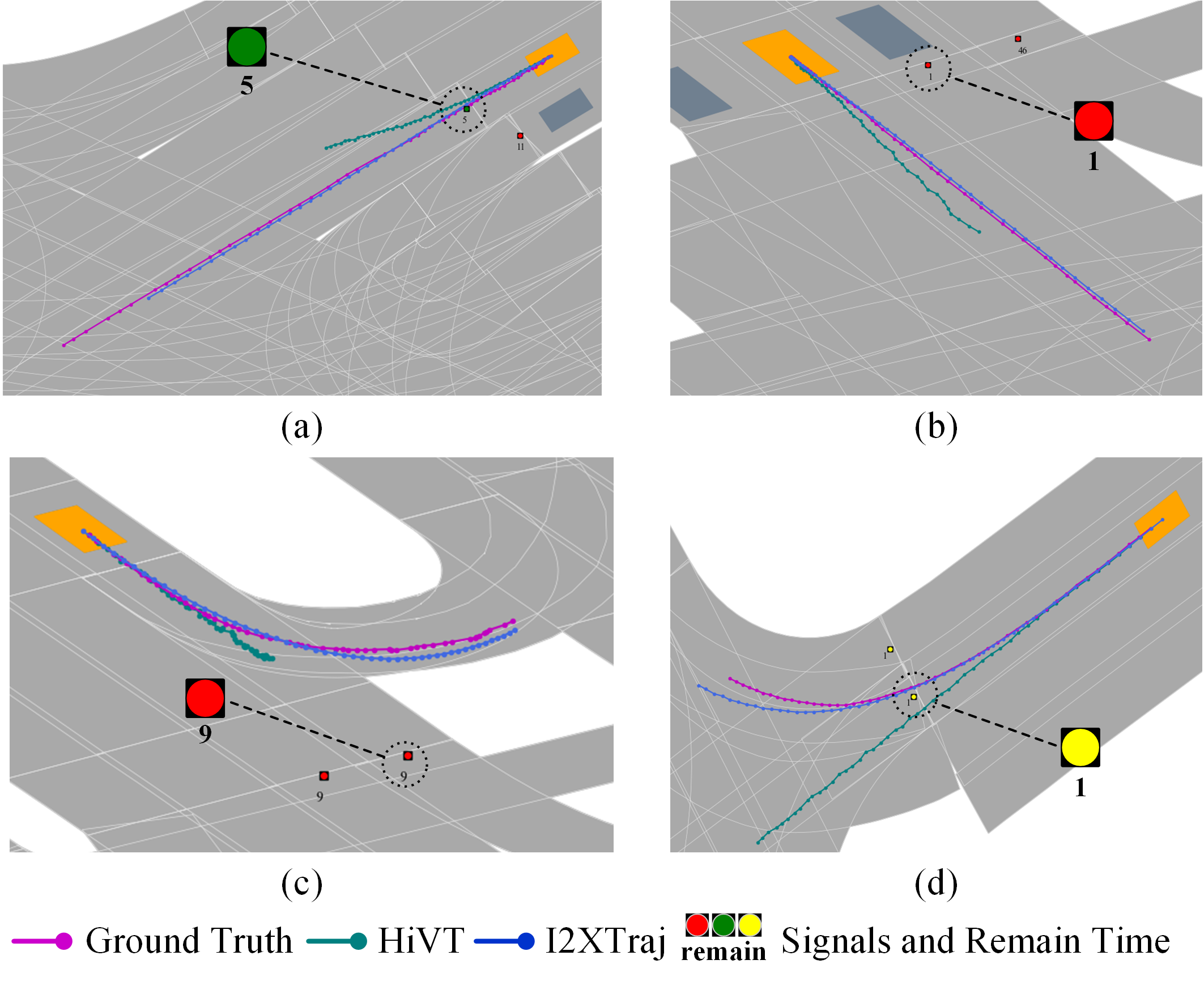}
\caption{Qualitative comparison of the influence of traffic signal lights. Target vehicles are depicted in orange, other vehicles are shown in gray. (a) and (b) illustrate the impact of traffic signals on vehicle speed strategies, while (c) and (d) show how signal control affects vehicle maneuvers in the regulated lane. Each figure shows the best trajectory among the $K$ modes.}
\label{fig: tl result}
\end{figure}

\subsection{Qualitative Results}
\subsubsection{Impact of Traffic Light Signals}
The impact of traffic signals on prediction results is difficult to quantify numerically. Through qualitative experiments, the influence of traffic signals on future trajectory can be observed intuitively. 

Fig. \ref{fig: tl result}(a) and (b) illustrate the impact of traffic signals on vehicle speed strategies at intersections.
Fig. \ref{fig: tl result}(a) demonstrates that when a green light is about to expire, vehicles tend to adopt an acceleration strategy to cross the intersection before the red light activates. In such scenarios, due to the initially low speed of the vehicle, a prediction model lacking traffic signal information can easily misinterpret and incorrectly predict deceleration or stop. Fig. \ref{fig: tl result}(b) illustrates how I2XTraj utilizes the final second of the red light signal to predict vehicle "jump-start" maneuver, wherein vehicles initiate movement prematurely during the red light phase and accelerate as the red light terminates.

\begin{figure*}[t]
\centering
\includegraphics[width=0.85\textwidth]{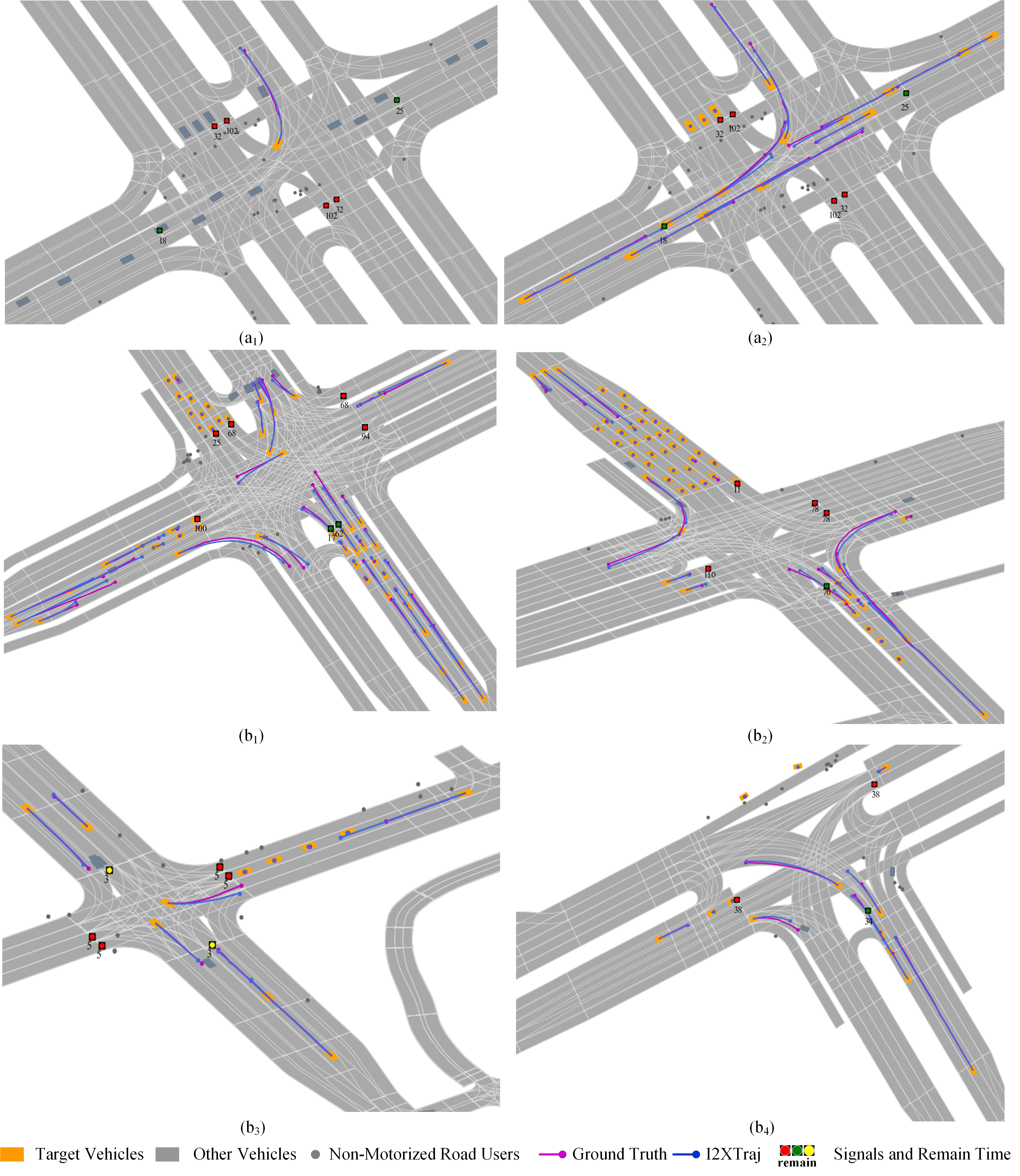}
\caption{Qualitative results of I2XTraj on six driving scenarios from V2X-Seq (SI) validation set. ($\text{a}_{1}$) and ($\text{a}_{2}$) illustrate single-agent and multi-agent prediction.
($\text{b}_{1}$), ($\text{b}_{2}$), ($\text{b}_{3}$), and ($\text{b}_{4}$) respectively present the results under four types of intersection scenarios, including large, medium, small and T-intersections. Each figure shows the best scene among the $K$ modes.}
\label{fig:Qualitative Results}
\end{figure*}

Fig. \ref{fig: tl result}(c) and (d) demonstrate vehicle maneuvering strategies on traffic signal-controlled lanes. 
In the scenario depicted in Fig. \ref{fig: tl result}(c), when a vehicle is traveling in a left-turn/U-turn lane, the left-turn maneuver is restricted by a red light. Rather than proceeding directly to the stop line to wait, the vehicle moves slowly before the dedicated U-turn lane, suggesting a high probability that the vehicle intends to execute a U-turn maneuver. Since I2XTraj comprehends that left turns are restricted through its analysis of traffic signals and lane control conditions, it successfully predicts the vehicle's U-turn maneuver.
Fig. \ref{fig: tl result}(d) illustrates that in a right-turn/straight lane, when the yellow light for proceeding straight illuminates and the vehicle has not yet decelerated to stop for the impending red light, the vehicle is more likely preparing to execute a right-turn maneuver. By leveraging the yellow signal information, I2XTraj effectively predicts the vehicle's right-turn trajectory.

\subsubsection{Qualitative Results in Different Scenarios}
The qualitative results demonstrate the trajectory prediction performance of I2XTraj across different scenarios.

Fig. \ref{fig:Qualitative Results}(a) compares single-agent and multi-agent prediction outcomes within the same scene. This comparison indicates that evaluating model performance based on a single target agent is reasonable and has been widely adopted. However, such an approach fails to capture the full complexity of intersection traffic. As shown in Fig. \ref{fig:Qualitative Results}(b), the global prediction involving multiple agents provides a comprehensive view. Jointly forecasting the trajectories of all vehicles at the intersection reflects the actual traffic dynamics in the scene.

Fig. \ref{fig:Qualitative Results}($\text{b}_\text{1}$), ($\text{b}_\text{2}$), ($\text{b}_\text{3}$) and ($\text{b}_\text{4}$)  demonstrate the generalization ability of I2XTraj across different types of intersections.
Fig. \ref{fig:Qualitative Results}($\text{b}_\text{1}$) presents the results at a large intersection with multiple bypasses. Fig. \ref{fig:Qualitative Results}($\text{b}_\text{2}$) shows the prediction performance at a typical medium-sized orthogonal intersection. Fig. \ref{fig:Qualitative Results}($\text{b}_\text{3}$) illustrates results at a small, narrow intersection shared by both pedestrians and cyclists. Fig. \ref{fig:Qualitative Results}($\text{b}_\text{4}$)  displays results at a T-shaped intersection. These scenarios indicate that I2XTraj can effectively predict the future trajectories of all vehicles across a wide range of intersection configurations. Notably, as shown in Fig. \ref{fig:Qualitative Results}($\text{b}_\text{1}$) and ($\text{b}_\text{2}$), the model maintains strong performance even in dense traffic conditions.

\subsection{Robustness Assessment}
\begin{table}
    \small
    \caption{Robustness experiment results of multi/single-agent on data loss}
    \label{tab:Robustness1}
    \centering
    \begin{tabular}{ccccc}
        \toprule
        Loss Ratio (\%)  & minADE  & minFDE & MR \\
        \midrule
        0        & 0.52/0.76    & 1.12/1.49  & 0.17/0.23    \\
        10       & 0.53/0.78    & 1.15/1.55  & 0.18/0.25    \\
        30       & 0.57/0.84    & 1.28/1.69  & 0.19/0.27    \\
        50       & 0.60/0.91    & 1.34/1.87  & 0.19/0.32    \\
        \bottomrule
    \end{tabular}
\end{table}
The I2X collaborative prediction scheme necessitates communication and data sharing among roadside sensors and RSUs. While this collaborative method provides comprehensive historical trajectory information for the prediction model, it also introduces common practical challenges in V2X communication delays, such as sensor failures or data loss, and communication delays. In real-world deployments, infrastructure communication equipment must transmit predicted trajectories to dozens or even hundreds of subscribing vehicles. In such multi-agent scenarios, the robustness of I2XTraj becomes particularly crucial.

\begin{table}
    \small
    \caption{Robustness experiment results of multi/single-agent on delays}
    \label{tab:delay}
    \centering
    \begin{tabular}{ccccc}
        \toprule
        Latency (ms)  & minADE  & minFDE & MR \\
        \midrule
        0         & 0.52/0.76      & 1.12/1.49  & 0.17/0.23   \\
        100       & 0.55/0.84      & 1.10/1.50  & 0.17/0.23    \\
        200       & 0.61/0.97      & 1.11/1.48  & 0.18/0.23    \\
        300       & 0.71/1.15      & 1.14/1.49  & 0.19/0.24    \\
        400       & 0.82/1.36      & 1.19/1.52  & 0.19/0.26    \\
        500       & 0.95/1.87      & 1.26/1.73  & 0.20/0.30    \\
        \bottomrule
    \end{tabular}
\end{table}

We evaluate the robustness to sensor failures or data loss by randomly discarding portions of historical data. In Table \ref{tab:Robustness1}, we tested the robustness under three typical data loss rates: 10\%, 30\%, and 50\%. The results demonstrate that even under extreme conditions with only half of the effective data available, I2XTraj still maintains viable prediction capability. More notably, under more common scenarios with minimal loss, less than 10\%, the performance in metrics remain at an exceptionally high level. 
This indicates that the model exhibits strong robustness against data loss. Comparing Table \ref{tab:Robustness1}, it is evident that the multi-agent approach demonstrates superior robustness compared to the single-agent approach, with lower performance degradation across all metrics. This phenomenon can primarily be attributed to the introduction of the strategy awareness mechanism. Vehicle dynamic information may become ambiguous due to data loss. At intersections, feasible maneuver strategies are governed by prior knowledge of traffic signals. Intersection geometry and prevailing traffic conditions further constrain these strategies. As a result, vehicles exhibit relatively stable selections of maneuver strategies. Extracting invariant maneuver strategies from changing dynamic features confers excellent robustness to the model. In scenarios considering all agents, the determination and prediction of feasible strategy combinations among agents further reduces the instability caused by data loss.

Furthermore, for cooperative systems, the delay robustness of the system presents an unavoidable challenge. The inference time of the prediction model and communication delays are compounded in the trajectory sequences that vehicles receive from the infrastructure. In general, the latency of a cooperative prediction system consists of two main components: the inference latency of the model and the communication transmission delay. The inference time of our model is 112ms, while actual communication conditions typically determine communication delays. Therefore, we conduct experiments on prediction results under conditions ranging from zero delay to 500ms delay to evaluate the delay robustness of the I2XTraj model in delayed environments in Table \ref{tab:delay}. Given that the model is designed for cloud-based or roadside edge computing units, this latency does not impede the real-time operation of autonomous driving systems while simultaneously reducing the computational burden on on-board platforms.
As shown in Table \ref{tab:delay}, I2XTraj demonstrates strong delay robustness. At a delay of 100ms, ADE, FDE, and MR exhibit only slight degradation, indicating that the performance decline caused by the inference time does not impact the superiority. The results from 100-500ms reveal that when communication delay does not exceed 400ms, the prediction results provided by the infrastructure remain reliable and effective.

Overall, quantitative analysis demonstrates that I2XTraj achieves outstanding performance in both single-agent and multi-agent trajectory prediction tasks. The incorporation of traffic light information and maneuvers priors significantly enhances prediction accuracy, particularly for maneuvering vehicles at intersections. Qualitative results further reveal that I2XTraj effectively handles diverse intersection scenarios with varying structural layouts and traffic densities. Finally, robustness experiments assess the impact of potential data loss and communication latency that may occur during real-world deployment.

\section{Conclusion}
\label{Conclusion}
In this work, we propose a trajectory prediction-centric I2X collaborative prediction scheme for signalized intersections. In this scheme, the infrastructure acquires comprehensive information and predicts the future trajectories of all vehicles therein. 
In the I2X collaborative prediction scheme, to address the unique characteristics of infrastructure-side data, we introduce I2XTraj, a prediction model specifically designed for deployment on infrastructure. I2XTraj is capable of jointly forecasting the future trajectories of all vehicles within an intersection.
Within I2XTraj, we incorporate a continuous signal-informed mechanism to capture real-time traffic control information. This mechanism establishes the dependency between traffic signals and future vehicle trajectories. In addition, a driving strategy awareness mechanism is proposed to model the joint distribution of vehicle maneuver strategies based on prior knowledge of typical intersections. This distribution guides the generation of joint scene trajectory proposals. Finally, a spatial-temporal-mode attention mechanism is employed to model interactions across three dimensions of the proposals, thereby refining predictions and enhancing trajectory accuracy. I2XTraj has demonstrated superior performance, surpassing SOTA methods by 30\% and 15\% on the V2X-Seq and SinD datasets, respectively. 
The qualitative results demonstrate the generalizability of our prediction model across diverse intersection geometries, while the robustness analysis establishes a solid foundation for its deployment in real‐world V2X environments.
The success of I2XTraj has suggested promising directions for future research in knowledge-driven trajectory prediction at signalized intersections.

It is important to note that this study primarily focuses on the infrastructure-side prediction model. Future work will explore how various autonomous driving systems, such as downstream planning and decision-making modules as well as end-to-end frameworks, can effectively leverage future trajectories provided by infrastructure to enhance their safety. Moreover, model lightweighting and efficiency optimization represent promising directions for future research.

\section*{Acknowledgment} 
The authors would like to thank TÜV SÜD for the kind and generous support. We are also grateful for the efforts of our colleagues in the Sino-German Center of Intelligent Systems. This work was also supported by the Fundamental Research Funds for the Central Universities.

\bibliography{reference.bib}
\bibliographystyle{IEEEtran}

\vfill

\end{document}